\documentclass[british,coneference]{IEEEtran}
\usepackage[T1]{fontenc}
\usepackage[latin9]{inputenc}
\usepackage{varwidth}
\usepackage{amstext}
\usepackage{graphicx}

\makeatletter

\providecommand{\tabularnewline}{\\}
\newenvironment{cellvarwidth}[1][t]
    {\begin{varwidth}[#1]{\linewidth}}
    {\@finalstrut\@arstrutbox\end{varwidth}}

\makeatother

\usepackage{babel}
\begin{document}
\title{Tractable Asymmetric Verification for Large Language Models via Deterministic
Replicability}
\author{%
\begin{minipage}[t]{0.3\textwidth}%
\begin{center}
Zan-Kai Chong\\
\emph{School of Science and\\ Technology\\
Kwansei Gakuin University}\\
Japan\\
zankai@ieee.org
\par\end{center}%
\end{minipage}%
\begin{minipage}[t]{0.3\textwidth}%
\begin{center}
Hiroyuki Ohsaki\\
\emph{School of Science and\\ Technology\\
Kwansei Gakuin University}\\
Japan\\
ohsaki@kwansei.ac.jp
\par\end{center}%
\end{minipage}%
\begin{minipage}[t]{0.3\textwidth}%
\begin{center}
Bryan Ng\\
\emph{ 
School of Engineering \& Computer Science\\
Victoria University of Wellington}\\
New Zealand
\\
ckbryan@hotmail.com
\par\end{center}%
\end{minipage}}
\maketitle
\begin{abstract}
The landscape of Large Language Models (LLMs) shifts rapidly towards
dynamic, multi-agent systems. This introduces a fundamental challenge
in establishing computational trust, specifically how one agent can
verify that another's output was genuinely produced by a claimed LLM,
and not falsified or generated by a cheaper or inferior model. To
address this challenge, this paper proposes a verification framework
that achieves tractable asymmetric effort, where the cost to verify
a computation is substantially lower than the cost to perform it.
Our approach is built upon the principle of deterministic replicability,
a property inherent to autoregressive models that strictly necessitates
a computationally homogeneous environment where all agents operate
on identical hardware and software stacks. Within this defined context,
our framework enables multiple validators to probabilistically audit
small, random segments of an LLM's output and it distributes the verification
workload effectively. The simulations demonstrated that targeted verification
can be over 12 times faster than full regeneration, with tunable parameters
to adjust the detection probability. By establishing a tractable mechanism
for auditable LLM systems, our work offers a foundational layer for
responsible AI and serves as a cornerstone for future research into
the more complex, heterogeneous multi-agent systems.

\end{abstract}

\section{Introduction\protect\label{sec:Introduction}}

The paradigm of large language models (LLMs) is rapidly shifting from
monolithic, single-turn systems to dynamic, multi-agent systems, where
autonomous agents collaborate to solve complex problems \cite{chen2025surveyllmbasedmultiagentsystem,luo2025largelanguagemodelagent}.
This evolution unlocks transformative potential in high-stakes domains
such as automated financial compliance, software engineering, and
scientific research, in which tasks can be delegated and synthesised
by specialised AI agents. This collaborative model, however, also
presents a fundamental challenge to the principles of responsible
AI: how can interacting agents reliably trust each other\textquoteright s
outputs as these systems are increasingly deployed in real-world applications?

This fundamental challenge arises when one agent generates an output
sequence, and another agent, acting independently, must determine
whether this output is the authentic result of the computation and
genuinely produced by the designated LLM under the specified conditions
(e.g., exact prompt, configuration, and constraints). Establishing
such provenance and integrity is essential in increasingly complex
and autonomous multi-agent workflows, where erroneous or unverified
outputs could propagate silently and have cascading, safety-critical
consequences, especially in regulated sectors such as finance and
healthcare \cite{tavasoli2025responsibleinnovationstrategicframework,reddy2023evaluating}.
For example, a recent discussion paper from Malaysia's central bank,
Bank Negara Malaysia (BNM), highlights the critical need for accountability
mechanisms to ensure responsible AI adoption in the financial sector
\cite{bnm2025dp_ai}. This regulatory focus implies a potentially
growing demand for crucial, auditable trails to ensure AI-generated
reports adhere to certified model standards and to enhance accountability
in the near future.

Addressing this challenge at scale requires a validation mechanism
built on the principle of asymmetric effort, also known as asymmetric
computation. This principle dictates that the computational cost required
to verify the correctness of a task must be significantly cheaper,
ideally by orders of magnitude lower than the cost to perform the
original task.

A canonical example of this principle is the Proof-of-Work (PoW) consensus
mechanism used in blockchain networks \cite{nakamoto2008bitcoin}.
A miner must expend a vast amount of computational energy (trillions
of hashes) to solve a difficult puzzle, representing the work. Once
a solution is found, any node on the network can verify its correctness
with a single, near-instantaneous calculation. Here, the effort is
highly asymmetric as the verification is exceptionally cheap compared
to the immense cost of the original computation. While our framework
does not use a blockchain, this principle of low-cost verification
is the bedrock of our approach to establish trust in multi-agent systems.

Central to this challenge is the lack of efficient, scalable mechanisms
for agents to verify the computational integrity of each other\textquoteright s
outputs in LLMs \cite{Ji_2023}. There is currently no direct method
to verify the output of a LLM with true asymmetric effort. Verifying
generative work naively requires re-running the same expensive inference
process. This cost is significant as the generation of output tokens
is priced substantially higher than the processing of input tokens
by commercial API providers, often by a factor of 3x to 5x, reflecting
the higher computational demand of generation \cite{Anthropic2025Pricing,OpenAI2025Pricing}.

To address this gap, we propose a validation framework that achieves
asymmetric effort through tractable mechanism. We imagine a scenario
with one agent acting as the generator and multiple agents as validators.
Our framework achieves an efficient, system-wide verification process
by leveraging the deterministic and autoregressive nature of LLMs
and probabilistic verification. Exploiting these properties, however,
introduces a critical requirement, i.e., a computationally homogeneous
environment. Although this confines the framework to controlled ecosystems,
it establishes a vital baseline for tractable low-cost verification
with asymmetrical effort.

The remainder of this paper is organised as follows. Section \ref{sec:Related-Work}
contextualises our contribution by reviewing related work in multi-agent
systems, LLM output verification, and the principles of verifiable
computation. Section \ref{sec:The-Verifiable-Task} details of our
proposed framework, presenting the core mechanisms of targeted validation
and probabilistic verification. Section \ref{sec:Experiment-Setup}
presents our experimental validation, where we quantify the tractable
asymmetric effort of our approach. Finally, Section \ref{sec:Conclusion}
concludes the paper and discusses avenues for future work.

\section{Related Work\protect\label{sec:Related-Work}}

Building on the principles of verifiable computation and asymmetric
effort discussed in Section \ref{sec:Introduction}, this section
positions our work within the existing academic landscape. We review
key literature across three relevant domains to highlight the specific
gaps our framework addresses. First, we explore trust and robustness
mechanisms in multi-agent systems. Second, we examine current approaches
to verifying LLM reasoning and outputs. Finally, we discuss the foundations
of verifiable computation to differentiate our framework from formal
cryptographic methods.

\subsection{Trust and Robustness in Multi-Agent Systems}

Multi-agent systems have traditionally confronted the challenge of
establishing trust and accountability among decentralised, heterogeneous
agents. Early solutions focused on reputation-based mechanisms \cite{sabater2004trust,sabater2005review},
while later work introduced game-theoretic incentives to foster cooperation
and reduce deception \cite{Pinyol2013}. The need for such mechanisms
is amplified in the high-stakes, regulated environments mentioned
earlier, where unverified agent actions can carry significant financial
or safety-critical consequences.

The emergence of LLM-based agents has introduced novel challenges.
These LLM agents often exhibit complex behaviours, i.e., unforeseen
patterns of interaction that arise from the interplay of multiple
agents. They can be difficult to interpret, predict, or control \cite{xu2025hallucinationinevitableinnatelimitation}.
Such dynamics are further explored in frameworks dedicated to emergent
multi-agent behaviours \cite{erisken2025maebemultiagentemergentbehavior}
and in studies documenting the spontaneous formation of social conventions
among LLM-agent populations \cite{ashery2025emergentsocialconventionscollective}.

A particularly acute issue is that LLM agents are prone to hallucinations,
i.e., generating outputs that are plausible and coherent, but factually
incorrect or misleading. Recent theoretical and empirical work shows
that hallucination is an inherent limitation of LLM architectures,
unavoidable in both single and multi-agent settings \cite{xu2025hallucinationinevitableinnatelimitation,karpowicz2025fundamentalimpossibilityhallucinationcontrol},
and surveys confirm that this challenge is pervasive across the field
\cite{Huang_2025}. Furthermore, these hallucinations can be expressed
with high certainty, making them especially difficult for traditional
trust mechanisms, such as confidence estimation or uncertainty quantification
to detect in real time \cite{simhi2025trustmeimwrong}.

Our work does not directly mitigate LLM hallucination but instead
focuses on establishing verifiable transparency. Specifically, our
method independently confirms whether an agent utilises the declared
input prompt and executes it with the specified parameters and LLM.
This process creates a crucial, auditable record of agent behaviour,
enabling stakeholders to trace the provenance of responses. While
this approach cannot eliminate hallucinations, it provides a reliable
foundation for accountability.

\subsection{Verification of LLM Reasoning and Outputs}

Alongside trust, the evaluation and verification of LLM outputs have
become a critical area of research. A popular evaluation methodology
is the concept of LLM-as-a-Judge, where a powerful LLM is used to
evaluate the output of another \cite{zheng2023judgingllmasajudgemtbenchchatbot}.
Typically, this involves a judge LLM scoring an output against a natural
language rubric. While useful for large-scale, automated evaluation,
this approach remains inherently subjective and focuses on the final
output rather than the process.

Recognising the limitations of purely outcome-based evaluation, researchers
have shifted focus to the reasoning process itself. A significant
step in this direction is the work by Cui et al. on verifiable misinformation
detection, which proposes an LLM agent that uses multiple tools to
generate an evidence log and a verifiable reasoning process \cite{cui2025verifiablemisinformationdetectionmultitool}.
Similarly, Yang et al. introduced CodeAgents, a framework that codifies
multi-agent reasoning into verifiable programs to improve planning
and token efficiency \cite{yang2025codeagentstokenefficientframeworkcodified}.
In a related vein, the analyst-inspector framework proposed by Nagarkar
et al. focuses on evaluating the reproducibility of LLM-generated
data science workflows \cite{zeng2025airepranalystinspectorframeworkevaluating}.

These approaches aim to make an agent's logical steps transparent
and replicable. Our work is complementary but distinct: instead of
verifying the correctness of the reasoning chain or the quality of
the final output, we focus on verifying the computational integrity
of each generative step. We provide a cryptographic-like guarantee
that an agent did, in fact, run the claimed model with the specified
inputs, a lower-level but fundamental layer of verification that underpins
the accountability of the entire reasoning process.

\subsection{Foundations in Verifiable Computation}

The principle of asymmetric effort is formally studied within the
field of verifiable computation. This principle enables a computationally
weak verifier to offload complex computations to a powerful but untrusted
prover. The verifier then receives a succinct, efficiently checkable
proof confirming that the computation was performed correctly \cite{verifiablecomputation2017}. 

A prominent class of solutions in this domain is Zero-Knowledge Proofs
(ZKPs). These cryptographic mechanisms allow a prover to demonstrate
that they have correctly executed a computation without revealing
any information about the inputs or intermediate steps.  However,
applying these powerful cryptographic techniques directly to the outputs
of LLMs presents formidable challenges. Current ZKP systems are primarily
designed for computations that can be expressed as algebraic circuits
(e.g., arithmetic operations). The core operations within a transformer
architecture, such as attention mechanisms and non-linear activation
functions, are not inherently crypto-native and are prohibitively
expensive to represent in this form \cite{sun2024zkllmzeroknowledgeproofs}.
Furthermore, the sheer scale of LLM inference, involving billions
or trillions of operations, makes the generation of such proofs computationally
infeasible with current methods.

Our work is inspired by the principles of verifiable computation but
in a different tractable approach. By tractable, we mean a framework
that is computationally feasible with current LLM architectures and
avoids the prohibitive overhead of formal methods like ZKPs. Instead
of seeking a cryptographic proof, we leverage the unique architectural
properties of LLMs to design a targeted and probabilistic verification
scheme that achieves asymmetric effort in a real-world setting.

\section{Tractable Asymmetric Verification \protect\label{sec:The-Verifiable-Task} }

This section details the framework for asymmetric verification proposed
in this paper. We begin by formalising the threat model that our system
is designed to mitigate in Section \ref{subsec:Threat-Model}. As
established in Section \ref{sec:Introduction} and \ref{sec:Related-Work},
achieving true asymmetric effort with LLMs is not directly feasible.
Our framework, therefore, implements two core mechanisms. First, in
Section \ref{subsec:targeted-validation}, we establish the foundation,
i.e., targeted validation, which leverages the deterministic replicability
of autoregressive models. Building on this, Section \ref{subsec:Distributed-Probabilistic-Verifi}
introduces distributed probabilistic verification, a multi-agent mechanism
that uses randomised sampling to achieve high collective security
with minimal computational overhead.

\subsection{Threat Model \protect\label{subsec:Threat-Model}}

Before detailing the proposed mechanism, it is essential to formalise
the threat model our framework is designed to address. We consider
an untrusted generator agent whose primary goal is to deceive other
agents within the system for strategic advantage. This adversarial
behaviour can manifest in two principal ways:
\begin{itemize}
\item Cost Evasion: The agent aims to reduce its computational expenditure
by substituting the output from the specified, computationally expensive
LLM with a response generated by a cheaper, inferior, or entirely
different model. In the most straightforward case, the entire output
sequence is fraudulent.
\item Malicious Content Injection: The agent seeks to subtly embed specific,
unverified, or harmful information within a larger, otherwise legitimate
output. In this scenario, the agent would only tamper with a small
fraction of the output segments to minimise the chances of detection.
\end{itemize}
Our framework is designed to be robust against both threats. While
a single random check would likely detect a full sequence replacement
stemming from cost evasion, the distributed probabilistic approach
is particularly crucial for countering malicious content injection.

\subsection{Deterministic Replicability in Targeted Validation\protect\label{subsec:targeted-validation}}

\begin{figure}
\begin{centering}
\includegraphics[width=0.95\columnwidth]{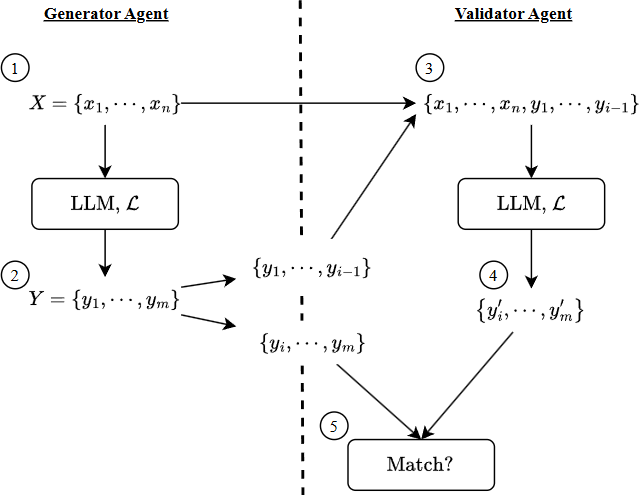}
\par\end{centering}
\caption{Verifying LLM generated content via deterministic replicability. (1)
The generator agent inputs prompt, $X$ into LLM $\mathcal{L}$. (2)
$\mathcal{L}$ generates the full output sequence, $Y$. (3) The validator
agent constructs a new input context by augmenting the prompt $X$
with a prefix of the original output of $Y$. (4) An identical LLM
instance generates new tokens from this context. (5) Finally, these
new tokens are compared with the corresponding tokens from the original
sequence $Y$ to verify a match. \protect\label{fig:generate-validate}}
\end{figure}

A foundational property of autoregressive LLMs is their deterministic
nature when used under controlled conditions, i.e., same seed number,
software stacks, GPU architecture, etc \cite{holtzman2020curiouscaseneuraltext,dodge2019workimprovedreportingexperimental,cui2025languagemodelsbayesianbrains}.
This section elucidates the principle that for a given model and a
fixed set of generation parameters, the output sequence is perfectly
replicable. 

Let an autoregressive LLM be denoted by $\mathcal{L}$. Given an initial
finite input sequence of tokens, $X=\left\{ x_{1},\dots,x_{n}\right\} $,
the model generates a finite output sequence $Y=\left\{ y_{1},\dots,y_{m}\right\} $,
where $n$ and $m$ denote the length of the input and output sequences
respectively. The generation process is sequential, where each token
$y_{i}$ is sampled from a conditional probability distribution computed
by the model: 
\begin{equation}
y_{i}\sim P\left(\cdot\,|\,x_{1},\dots,x_{n},y_{1,\dots},y_{i-1}\right).
\end{equation}

The central proposition is that this generation process is computationally
deterministic. If we construct a new prompt $X^{\prime}$ by concatenating
the original prompt with the first $j$ tokens of its output, where
$j<m$, such that $X^{\prime}=\left\{ x_{1},\dots,x_{n},y_{1,\dots},y_{j}\right\} $,
$\mathcal{L}$ will subsequently generate the exact remaining sequence
$Y^{\prime}=\left\{ y_{j+1},y_{j+2},\dots,y_{m}\right\} \subseteq Y$,
provided all external conditions are held constant.

This property provides a computationally efficient mechanism for a
validator agent to verify the authenticity of a sequence generated
by a generator agent. Instead of regenerating the entire output sequence
$Y$, the validator agent can perform targeted validation, as illustrated
in Figure \ref{fig:generate-validate}. To validate any specific token
$y_{j}$ within the sequence, the validator agent only needs to provide
the preceding input context $\left\{ x_{1},\dots,x_{n},y_{1},\dots,y_{j-1}\right\} $
to an identical instance of $\mathcal{L}$ with the same parameters,
then it can regenerate the subsequent sequence --- from a single
token to the entire remainder. This mechanism enables robust, trustless
verification at a fraction of the original cost, potentially satisfying
the principle of asymmetric effort.

\subsection{Distributed Probabilistic Verification\protect\label{subsec:Distributed-Probabilistic-Verifi}}

\begin{figure}
\begin{centering}
\includegraphics[width=0.5\columnwidth]{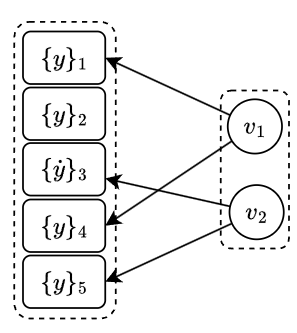}\caption{An illustration of the distributed probabilistic verification mechanism.
An output is divided into five segments, with $\left\{ \dot{y}\right\} _{3}$
being tampered. Two validators, $v_{1}$ and $v_{2}$, independently
sample segments to verify. The mechanism succeeds as validator $v_{2}$
selects the tampered segment $\left\{ y\right\} _{3}$ in the verification
process. \protect\label{fig:distributed-proba-verification}}
\par\end{centering}
\end{figure}

While the deterministic replicability detailed in Section \ref{subsec:targeted-validation}
offers a robust validation mechanism, it presents a practical challenge:
unless the tampered position is predictable, an exhaustive, token-by-token
search remains necessary. This computationally intractable requirement
undermines the asymmetric effort that the approach seeks to establish.

To overcome this, our framework leverages the multi-agent context
to transform this search problem into an efficient probabilistic sampling
task. In a system with a single generator and multiple validator agents,
the verification burden can be distributed. Instead of a single agent
validating the entire sequence, each validator is tasked with checking
only a small, randomly selected segment of the output.

The mechanism for distributed probabilistic verification is formalised
as follows. Let the generator's complete output, $Y$, be conceptually
divided into $k$ verifiable segments, i.e., $Y=\left\{ \left\{ y\right\} _{1},\cdots,\left\{ y\right\} _{k}\right\} $.
Without loss of generality, we assume each segment of equal size.
Suppose a malicious agent modifies a subset of these segments, creating
$f$ tampered segments i.e., $\left\{ \left\{ \dot{y}\right\} _{1},\cdots,\left\{ \dot{y}\right\} _{f}\right\} $,
where $1\leq f\ll k$. Additionally, the system contains $q$ independent
validator agents, each tasked with verifying $r$ distinct segment
($r\geq1$), chosen uniformly at random without replacement. This
mechanism is illustrated in Figure \ref{fig:distributed-proba-verification}.

For a single validator to fail, all $r$ segments it selects must
be from the $k-f$ valid segments. The total number of ways to choose
$r$ segments from $k$ is given by the binomial coefficient ${k \choose r}$.
The number of ways to choose $r$ segments exclusively from the valid
pool is ${k-f \choose r}$. Denote the probability that a single validator
fails to detect the tampered segments as $\text{P}\left(\text{Single Fail}\right)$
and this is defined as,
\begin{equation}
\text{P}\left(\text{Single Fail}\right)=\frac{{k-f \choose r}}{{k \choose r}}.
\end{equation}

Since each of the $q$ validators makes their selection independently,
the probability that all of them fail to detect the tampered segments
is the product of their individual failure probabilities. Consequently,
the probability of the mechanism's success is the complement of the
failure of $q$ validators (denoted by $\text{P}\left(\text{Detect}\right)$),
i.e.,
\begin{equation}
\text{P}\left(\text{Detect}\right)=1-\left(\frac{{k-f \choose r}}{{k \choose r}}\right)^{q}.
\end{equation}

This mathematical relationship demonstrates the power of the distributed
approach. The system's security now grows not only exponentially with
the number of validators, $q$, but also increases significantly with
the number of segments each validator checks, $r$. This provides
a highly tunable framework for achieving a desired level of security,
offering a robust and scalable foundation for collective trust.

While a detailed communication protocol is beyond the scope of this
paper, the framework assumes that if any validator detects a mismatch,
this finding can be broadcast to other agents to trigger a consensus-based
rejection of the fraudulent output. This ensures that system-wide
integrity can be maintained with minimal computational overhead for
each participating agent. 

\section{Experiment Setup\protect\label{sec:Experiment-Setup}}

To validate the claims of our proposed framework, we conducted a series
of simulations designed to quantify its performance and efficiency.
The experiments were conducted using Llama 3 from HuggingFace \cite{meta_llama_3_8b_instruct_2024},
a widely accessible open-source autoregressive model. All simulations
were run on a single NVIDIA RTX 4000 Ada GPU, using a standard PyTorch
and Hugging Face transformers implementation. We also discuss the
critical impact of hardware heterogeneity on deterministic replicability.

\subsection{Quantifying Asymmetric Effort}

Our first experiment aimed to quantify the core principle of asymmetric
effort. We measured the computational cost of generating a complete
sequence and compared it against the cost of performing targeted validation
on segments of varying lengths. To elicit a long, sequential output
suitable for this test, we used a two-part Chain-of-Thought (CoT)
financial reasoning problem, which resulted in a 792-token response.
The primary metric was wall-clock time, measured in seconds.

The results presented in Table \ref{tab:Generation-cost} demonstrate
a significant disparity between the cost of full generation and targeted
validation. The full generation required 32.13 seconds to complete.
In stark contrast, verifying the final 50 tokens of the same sequence
took only 2.59 seconds, representing an asymmetric effort ratio of
12.41x, which is calculated by dividing the full generation time by
the verification time.

As anticipated, the verification time scales linearly with the length
of the segment being validated. While the asymmetry is most pronounced
for smaller segments, even verifying a substantial portion of the
output (200 tokens) is over three times faster than a full re-generation.
These findings empirically validate the foundational premise of our
framework, i.e., targeted validation provides a substantial reduction
in computational cost, making it a feasible mechanism for establishing
trust in multi-agent systems.

As noted in Section \ref{subsec:targeted-validation}, the output
of a targeted validation will always align with the corresponding
portion of the original sequence in the controlled conditions. Hence,
we will not revisit it here. Furthermore, verification can be performed
from any arbitrary position rather than only generating the final
tokens by appropriately controlling the input context and the number
of tokens generated for validation.

\begin{table}
\caption{Generation cost vs. targeted validation cost. \protect\label{tab:Generation-cost}}

\begin{centering}
\begin{tabular}{|c|c|c|}
\hline 
Generation & \begin{cellvarwidth}[t]
\centering
Verification 

Time (s)
\end{cellvarwidth} & \begin{cellvarwidth}[t]
\centering
Asymmetric 

Effort Ratio
\end{cellvarwidth}\tabularnewline
\hline 
\hline 
Full 792 tokens & 32.13 & ---\tabularnewline
\hline 
Last 50 tokens & 2.59 & 12.41\tabularnewline
\hline 
Last 100 tokens & 5.25 & 6.12\tabularnewline
\hline 
Last 200 tokens & 10.01 & 3.21\tabularnewline
\hline 
Last 400 tokens & 19.85 & 1.62\tabularnewline
\hline 
\end{tabular}
\par\end{centering}
\end{table}

\subsection{Efficacy of Distributed Probabilistic Verification}

\begin{figure}

\begin{centering}
\includegraphics[width=0.9\columnwidth]{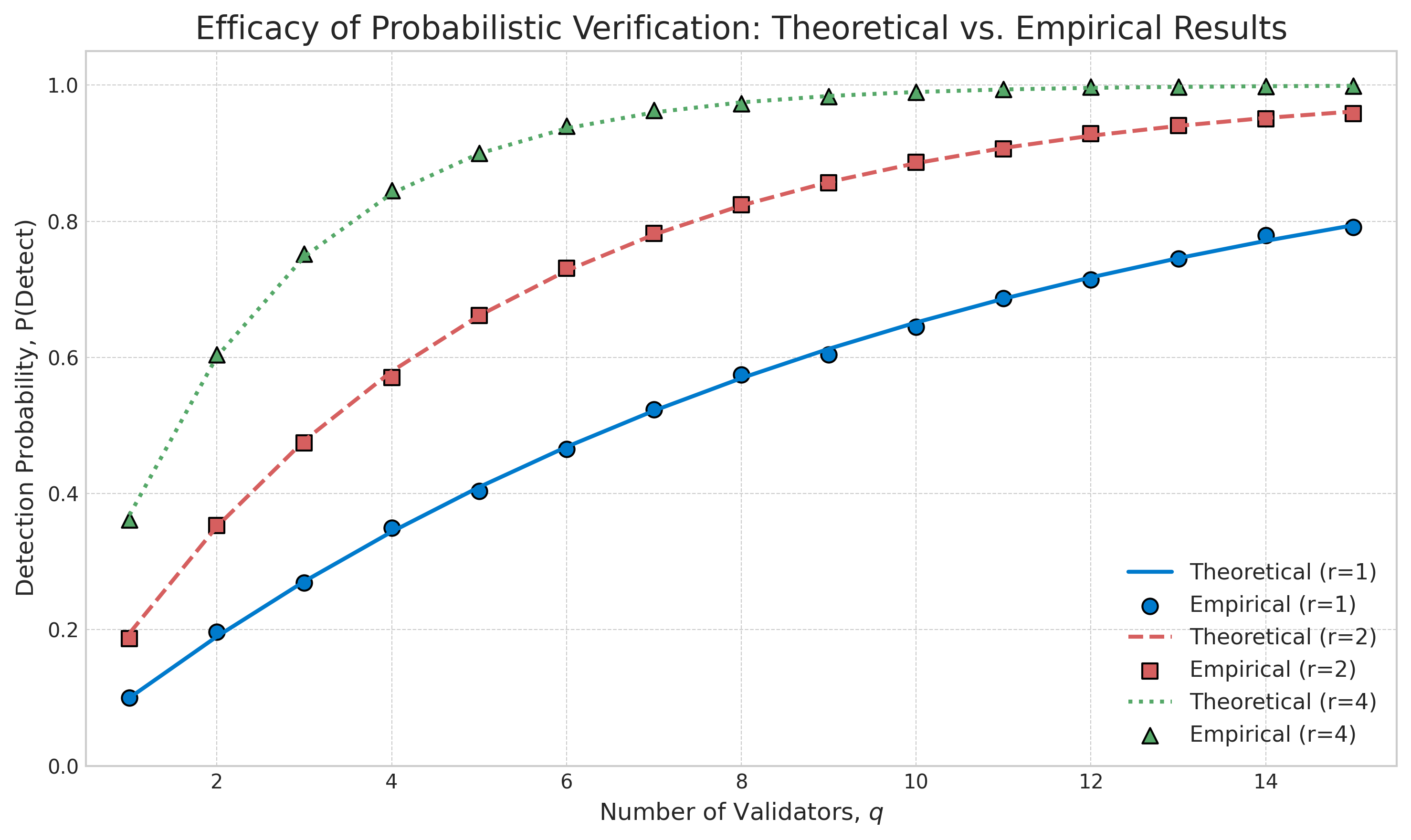}
\par\end{centering}
\caption{Validation of the probabilistic verification with $k=20$ segments
and $f=2$ fraudulent. The lines represent the theoretical detection
probability for different numbers of checks per validator, $r$, while
markers show the empirical results from 10,000 trials per point. The
close alignment confirms the model's predictive accuracy. \protect\label{fig:proba-veri}}
\end{figure}

Our second experiment was designed to validate the mathematical model
of the verification's efficacy presented in Section \ref{subsec:Distributed-Probabilistic-Verifi}.
The objective was to demonstrate that the detection probability scales
predictably with the number of validators, $q$ and the number of
segments each validator checks, $r$.

To achieve this, we simulated a scenario where a malicious agent tampered
with a generated output. The output was defined as having a total
of $k=20$ segments, with $f=2$ of those segments being tampered
(a 10\% tamper rate). We then compared the theoretical detection probability
calculated directly from our formula against the empirical results
from a numerical simulation of the random sampling process. Each empirical
data point was generated by running 10,000 independent simulation
trials.

The results are presented in Figure \ref{fig:proba-veri} with the
solid and dashed lines represent the theoretical detection probability,
while the markers represent the empirical results from the simulation.
As shown, the empirical data points align almost perfectly with the
theoretical curves for $r=1$ to 4. This close alignment provides
strong validation for our mathematical model, confirming that it accurately
predicts the tractable performance of the verification mechanism.

The findings illustrate the highly tuneable nature of the framework's
security. For instance, with each validator checking only one segment
($r=1$), increasing the number of validators from 5 to 10 boosts
the detection probability from approximately 40\% to 65\%. Furthermore,
by increasing the workload slightly to have each of the 10 validators
check two segments ($r=2$), the detection probability rises dramatically
to over 88\%. This experiment confirms that the distributed probabilistic
verification method provides a reliable and predictable means of achieving
a desired level of collective security.

\subsection{Discussion on Determinism and Hardware Constraints}

A core assumption of our framework is the deterministic replicability
of LLM outputs under controlled conditions. To investigate the boundaries
of this assumption, we conducted preliminary tests to assess the impact
of heterogeneous hardware on verification. Our tests confirmed that
generating an output sequence on one GPU model (NVIDIA RTX 4000 Ada)
and attempting to validate it on a different GPU model (NVIDIA A40)
led to verification failures. These failures arise from minute, non-malicious
floating-point discrepancies inherent in different hardware architectures.

This finding establishes that a practical implementation of our verification
system requires a strict, homogeneous computational environment. All
participating agents, both generators and validators, must operate
using not only the same model and configuration parameters but also
identical hardware and software stacks. Therefore, we posit that a
standardised execution environment, such as a specific cloud GPU instance
or a strictly version-controlled container, is a prerequisite for
deploying this framework in a real-world decentralized network. Addressing
this hardware-induced non-determinism, perhaps through fuzzy verification
methods, remains a key area for future work.

\section{Conclusion\protect\label{sec:Conclusion}}

In this paper, we addressed the critical challenge of ensuring computational
integrity for LLMs within multi-agent systems, where the lack of a
tractable, low-cost verification mechanism is a significant barrier
to establishing trust and provenance. To bridge this gap, we proposed
a framework that achieves tractable asymmetric effort by combining
targeted validation and distributed probabilistic verification. Our
simulations empirically validated this approach, demonstrating a significant
reduction in computational cost and confirming the predictive accuracy
of the framework. While our work provides a robust foundation for
building auditable systems, we also experimentally confirmed that
strict hardware and software homogeneity is a critical prerequisite
for verification, as different GPU models produce non-identical outputs.
This finding highlights that crucial avenues for future work include
developing standardised execution environments and verification techniques
resilient to these minor, hardware-induced discrepancies.

\bibliographystyle{IEEEtran}
\bibliography{main}

\end{document}